\documentclass[conference]{IEEEtran}
\IEEEoverridecommandlockouts 

\usepackage{cite}
\usepackage{amsmath,amssymb,amsfonts}
\usepackage{algorithmic}
\usepackage{textcomp}
\usepackage{xcolor}
\usepackage{times}
\usepackage{epsfig}
\usepackage{graphicx} 
\usepackage{multirow}
\usepackage{subfigure}
\usepackage{pgfplots}
\usepackage{lettrine}
\usepackage[pagebackref,breaklinks=true,bookmarks=false,hidelinks=false,colorlinks,linkcolor=red,anchorcolor=blue,citecolor=green]{hyperref}
\def\BibTeX{{\rm B\kern-.05em{\sc i\kern-.025em b}\kern-.08em
    T\kern-.1667em\lower.7ex\hbox{E}\kern-.125emX}}

\begin{document}

\title{Blind Predicting Similar Quality Map for Image Quality Assessment
}

\author{\IEEEauthorblockN{1\textsuperscript{st} Da Pan, 2\textsuperscript{nd}Ping Shi , 3\textsuperscript{rd} Ming Hou, 4\textsuperscript{th}  Zefeng Ying, 5\textsuperscript{th} Sizhe Fu, 6\textsuperscript{th} Yuan Zhang}
\IEEEauthorblockA{\textit{Communication University of China} \\
No.1 Dingfuzhuang East Street Chaoyang District, Beijing, China \\
\{pdmeng, yingzf, shiping, houming, fusizhe, yzhang\}@cuc.edu.cn}}
\maketitle
\begin{abstract}
A key problem in blind image quality assessment (BIQA) is how to effectively model the properties of human visual system in a data-driven manner. In this paper, we propose a simple and efficient BIQA model based on a novel framework which consists of a fully convolutional neural network (FCNN) and a pooling network to solve this problem. In principle, FCNN is capable of predicting a pixel-by-pixel similar quality map only from a distorted image by using the intermediate similarity maps derived from conventional full-reference image quality assessment methods. The predicted pixel-by-pixel quality maps have good consistency with the distortion correlations between the reference and distorted images. Finally, a deep pooling network regresses the quality map into a score. Experiments have demonstrated that our predictions outperform many state-of-the-art BIQA methods.
\end{abstract}

\section{Introduction}
\begin{figure}[t]
\begin{center}
\subfigure[]{
  \begin{minipage}{4cm}
  \includegraphics[width=4.0cm,height=2.8cm]{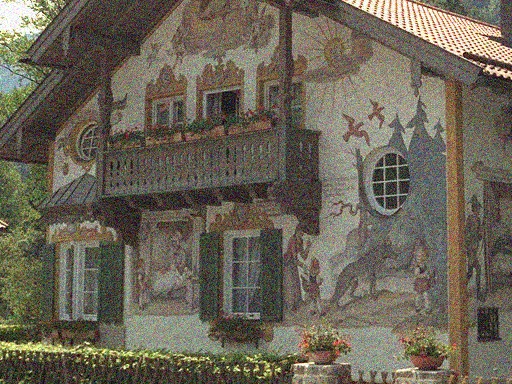}\vspace{0.25cm}\label{Fig1:a}
\end{minipage}}
\subfigure[]{
  \begin{minipage}{4cm}
  \includegraphics[width=4.0cm,height=2.8cm]{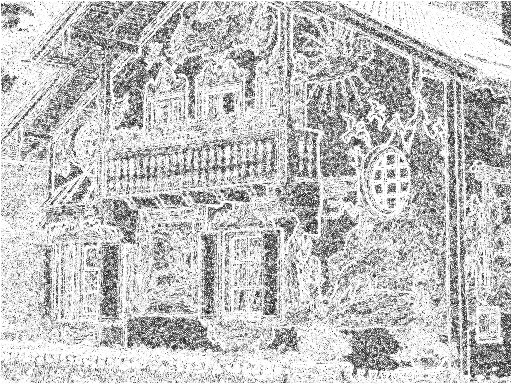}\vspace{0.25cm}\label{Fig1:b}
\end{minipage}}
\vfill
\subfigure[]{
  \begin{minipage}{4cm}
  \includegraphics[width=4.0cm,height=2.8cm]{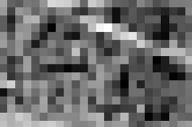}\vspace{0.25cm}\label{Fig1:c}
\end{minipage}}
\subfigure[]{
  \begin{minipage}{4cm}
  \includegraphics[width=4.0cm,height=2.8cm]{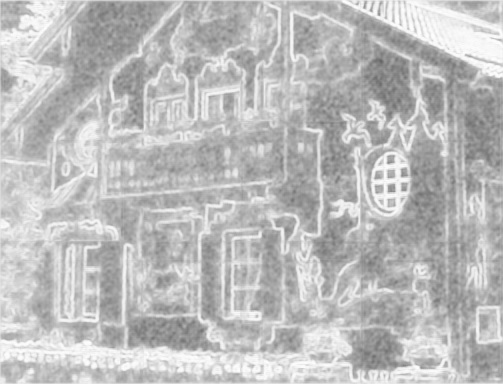}\vspace{0.25cm}\label{Fig1:d}
\end{minipage}}
 \caption{Examples of predicted quality maps: (a) is a distorted image; (b) is a similarity map from FSIM; (c) is a patch-based quality map from BIECON ~\cite{kimfully}; (d) is a pixel-based quality map predicted from our proposed model.}\label{fig:1} 
\vspace{-0.75cm}
\end{center}
\end{figure}

\lettrine[lines=2]{\textbf{O}}{\textbf{bjective}} image quality assessment (IQA) is a fundamental problem in computer vision and plays an important role in monitoring image quality degradations, optimizing image processing systems and improving video encoding algorithms. Therefore, it is of great significance to build an accurate IQA model. In the literature, some full-reference image quality assessment (FR-IQA) methods~\cite{4--1,fsim-15,GMSD--1,MDSI1,DOG1,harr,VSI} which attempt to build a model simulating human visual system (HVS) can achieve good performance. For example, FSIM~\cite{fsim-15} predicts a single quality score from a generative similarity map (as shown in Fig. \ref{fig:1}(b)). According to our analysis, two reasons bring FR-IQA methods into success. One reason is that it can access to reference image content and take the reference information by comparison. Meanwhile, this way of comparison is similar with the behavior of human vision and makes it easy to judge the image quality by FR-IQA methods~\cite{ss}. The other reason is that hand-crafted features carefully designed by FR-IQA are closely related to some HVS methods properties. The difference of features on corresponding positions between reference and distorted images can well measure the distortion degree. On the other hand, some NR-IQA methods~\cite{DIIVINE-18,no--3,100,NIQE-21} which rely on natural scene statistics do not obtain the same satisfying performance. As a result, the accuracies of most FR-IQA methods are better than those of NR-IQA when the performance is objectively evaluated. 

Based on these analysis, it is difficult for NR-IQA methods to build a model to imitate the behavior of HVS under the case of lacking reference information. Recently, researchers have started to harness the power of convolutional neural networks (CNNs) to learn discriminative features for various distortions types~\cite{pa-1,Live14,kang27,w-nr-fr}. We name these methods Deep-IQA. Most previous Deep-IQA methods just consider CNN as a complicated regression function or a feature extractor, but are unaware of the importance of generating intermediate quality maps which represent the perceptual impact of image quality degradations. This training process of Deep-IQA seems not to have an explicit perceptual meaning and is always a black box for researchers. But what interests us is that BIECON~\cite{kimfully} proposed an idea that training a CNN to replicate a conventional FR-IQA such as SSIM~\cite{4--1} or FSIM~\cite{fsim-15}. However, the method estimates each local patch score and patch-wise scores are pooled to an overall quality score. In essence, it visualizes a score patch map which contains spatial distribution information and the map does not reflect the distorted image in pixel level, as shown in Fig.~\ref{fig:1}(c). But we consider that the distortion value of each pixel is affected by its neighboring pixels and should not be exactly the same in the same patch. The simple patch-based scheme is not enough to correlate well with perceived quality. Therefore, how to design an effective deep learning model for blind predicting an overall pixel-by-pixel quality map related to human vision is the focus of this work.

In this paper, we propose a new Deep-IQA model which consists of a fully convolutional neural network (FCNN) and a deep pooling network (DPN). We refer to this method as Blind Predicting Similar Quality Map for IQA (BPSQM). Specifically, given a similarity index map label, our proposed model can produce a HVS-related quality map to approach to the similarity index map in pixel distortion level. The predicted quality map can be a measurement map for describing the distorted image. Intuitively, the FCNN tries to simulate the process of FR-IQA methods generating similarity index maps. Then, given a subjective score label, the DPN which can be equivalent to various complicated pooling strategies predicts a global image quality score based on the predicted quality map. The primary advantage for this model is that the additional similarity map label guides FCNN to learn local pixel distortion features in the intermediate layers. Our proposed model considers assessing image quality as a problem of image-to-image. The quality maps predicted from BPSQM can reflect distorted areas in pixel level. Meanwhile, our model is simple and effective.

Our key insight is that good guided learning policies can help NR-IQA methods accurately predict global similar quality maps which agree with the distortion distribution between reference and distorted images. We use HVS-related similarity index maps derived from FR-IQA methods to navigate the learning direction of FCNN. Through guided learning, FR-IQA methods can transmit HVS-related pixel distortion feature information to NR-IQA methods. Fig. ~\ref{fig:1}(d) shows a generative quality map from BPSQM. Compared to the patch-based quality map in (c), it is obvious that (d) represents pixel-wise distortions for a global distorted image. Meanwhile, the distortion distribution is generally similar with the feature map (b) from FSIM. In addition, a deep pooling network used for predicting the perceptual image quality is superior to other pooling strategies.

This paper has the following innovations and highlights.

(1) Numerous experiments indicate that a similarity quality map in pixel distortion level can achieve good consistency with human perception by feeding it into a learnable pooling network, which is our discovery and becomes the starting point of our paper. 

(2) This paper firstly introduces the pixel-to-pixel segmentation method into IQA to predict global similarity quality maps in pixel level to be close to quality maps from FR-IQA.

(3) Our predicted quality maps can provide us an intuitive analysis of local distortions, which enables to improve image enhancement module.

The rest of the paper is structured as follows. Section II introduces related work including FR-IQA, NR-IQA and Deep-IQA methods. Section III shows the detail architecture of the proposed BPSQM framework. The important cornerstone experiment and other extensive comparison experiments are given in Section IV. We conclude the paper in Section V.

\section{Related Work}

\subsection{Full-reference Image Quality Assessment} 

\begin{figure*}[pt]   
  \centering
  \includegraphics[scale=0.6]{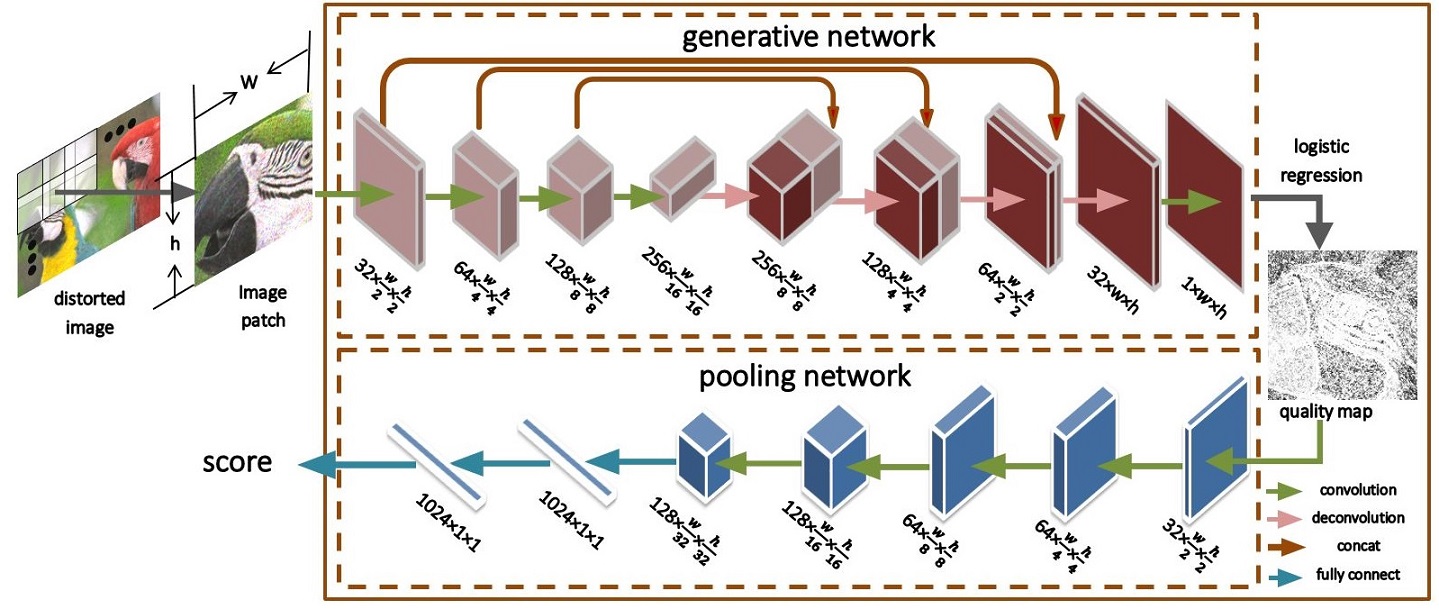}\\
  \caption{Architecture of the proposed BPSQM framework. The generative network takes as input a distorted image and predicts a similar quality map related to human vision. The pooling network directly regresses the generative quality map into a score.}\label{Network}
\end{figure*}

In order to effectively model the properties of HVS, many HVS-related methods have been proposed. The structure similarity index (SSIM) \cite{4--1} extracted the structural, contrast and luminance information to constitute a similarity index map for assessing the perceived image quality. In~\cite{fsim-15}, Zhang ,et al. proposed a feature-similarity index which calculated the phase congruency (PC) and gradient magnitude (GM) as features for the HVS perception.~\cite{GMSD--1} proposed an efficient and effective standard deviation pooling strategy, which demonstrates that the image gradient magnitude alone can still achieve high consistency with the subjective evaluations.~\cite{MDSI1} used a novel deviation pooling to compute the quality score from the new gradient and chromaticity similarities, which further suggests that the gradient similarity could well measure local structural distortions. The aforementioned FR-IQA methods first compute a similarity index map to represent some properties of HVS and then design a simple pooling strategy to convert the map into a single quality score.

\subsection{No-reference Image Quality Assessment} 

Many NR-IQA approaches model statistics of natural images and exploit parametric variation from this model to estimate perceived quality. DIIVINE~\cite{DIIVINE-18} framework identified the distortion type firstly and applied a distortion-specific regression strategy to predict image quality degradations. BLIINDS-II~\cite{bok-1} presented a Bayesian inference model to give image quality scores based on a statistical model of discrete cosine transform (DCT) coefficients. The CORNIA~\cite{100} learned a dictionary from a set of unlabeled raw image patches to encode features, and then adopted a max pooling scheme to predict distorted image quality. NIQE~\cite{NIQE-21} used a multivariate Gaussian model to obtain features which are used to predict perceived quality in an unsupervised manner. SOM~\cite{SOM-25} focused on areas with obvious semantic information, where the patches from the object-like regions were input to CORNIA.

\subsection{Deep Image Quality Assessment}  

With the rise of CNN for detection and segmentation tasks~\cite{kai1,kai2,kai3}, more and more researchers have started to apply the deep network into IQA. Lu et al.~\cite{deep38} proposed a multi-patch aggregation network based on CNN, which integrates shared feature learning and aggregation function learning. Kang et al.~\cite{kang27} constructed a shallow CNN only consisting of one convolutional layer to predict subjective scores.~\cite{w-nr-fr} proposed a deeper network with 10 convolutional layers for IQA.~\cite{iccv2017} employed ResNet~\cite{resnet} to extract high-level features to reflect hierarchical degradation. Most Deep-IQA methods only employ CNN to extract discriminative features and are inadequate for analyzing and visualizing the intermediate results, which makes it difficult for us to understand how to process IQA based on CNN. In~\cite{cvpr2017}, Kim et al. proposed a full reference Deep-IQA generating a perceptual error map which provides us an intuitive analysis of local artifacts for given distorted images. BIECON~\cite{kimfully} designed a deep network to  estimate a patch score map and utilized one hidden layer to regress the extracted patch-wise features into a subjective score.

\section{Quality Map Prediction}  

\textbf{Problem formulation}

Given a color or gray distorted image \emph{$I_{d}$}, our goal is to estimate its quality score by modeling image distortions. Previous works for deep NR-IQA~\cite{kang27,w-nr-fr} lets \emph{$f_{\theta}$} be a regression function using CNN with parameter $\theta$. \emph{S} indicates the subjective ground-truth score:
\begin{equation}\label{equ1}
  \centering
  S = f_{\theta}(I_{d})
\end{equation}

In this case, the deep network simply trains on input images and directly outputs results. From the process, we can not understand how the deep network learns features related with the image distortion. In contrast, FR-IQA methods generate similarity index map firstly and then pool the map. The process can be formulated as below:
\begin{equation}\label{equ2}
  \centering
  S = P(M(I_{d},I_{r}))
\end{equation}
Where \emph{$I_{r}$} represents a reference image. \emph{M} indicates the way to calculate similarity index map and \emph{P} denotes a pooling strategy, for example, \emph{P} can be a simple average operation in SSIM or a standard deviation operation in GMSD~\cite{Xue2014Gradient}. Given all that, we combine the advantage of FR-IQA modeling general properties of HVS with the advantage of NR Deep-IQA without hand-crafted features. Our approach firstly constructs a generative network \emph{G} with parameters $\omega$ to predict a global quality map in pixel level. Then, a deep pooling network $f_{\phi}$ regarded as a complicated pooling strategy converts the predicted quality map into a score.

\begin{equation}\label{equ2}
  \centering
  S = f_{\phi}(G_{\omega}(I_{d}))
\end{equation}
\subsection{Architecture} 

The proposed overall framework is illustrated in Fig.~\ref{Network}. This framework consists of two main components: a generative quality map network and a quality pooling network. The requirement for the generative network is to output a quality map of the same size with the input image. We select U-Net~\cite{unet}, an extension of FCNN, as a base of generative network. Because U-Net integrates the hierarchical representations in subsampling layers with the corresponding features in upsampling layers. So the degradations on both the low and high level features are considered for IQA~\cite{iccv2017}.

The generative network consists of a subsampling path (SP) and an upsampling path (UP). In the SP, the distorted image goes through four convolutional layers with kernel 3$\times$3 and padding 1$\times$1. In the UP, there are also four corresponding deconvolution layers with kernel 2$\times$2 and stride 2$\times$2. The feature maps in the SP are contacted with the corresponding feature maps of the same size in UP. The last deconvolution layer outputs a pixel-wise dense prediction map with the same size as the input image. Batch normalization~\cite{BN} and leaky rectified linear unit (LReLU) are used after all convolution and deconvolution operations. A 3$\times$3 convolutional layer with padding 1$\times$1 for keeping same size is used for reducing dimensionality into one channel feature map. The feature map is input into a sigmoid function and squashed into the [0,1], the loss function is binary cross entropy loss.\footnote{In MxNet framework, the procedure can be implemented with a LogisticRegressionOutput layer.} In our paper, the pooling network contains five 3$\times$3 convolutional layers with 2$\times$2 maximum pooling and two fully connected layers. We perform 50\% dropout after each fully connected layers so as to prevent overfitting. The pooling network ends up with a squared Euclidean loss layer. It should be noted that we crop the input image into some overlapping fixed size patches so as to adapt to the pooling network. This patch size should be large enough, which will not influence the learning of pixel distortion.

\subsection{Quality Map Selection} 

SSIM~\cite{4--1}, FSIM~\cite{fsim-15} and MDSI~\cite{MDSI1} are adopted to generate similarity maps as label separately. Because the luminance, contrast and structural information are treated equally in SSIM, the similarity map derived from SSIM is directly used as map label. In contrast, the FSIM method uses a pooling weight to combine the phase congruency (PC) and gradient magnitude (GM) in computing the final quality score. So we select the two features as map label separately. As for MDSI~\cite{MDSI1}, the combination of gradient and chromaticity similarity maps is selected as label.

We remove pre-processing including filtering and downsampling in the process of computing the similarity index map label to guarantee the generative map same size with the input image. Specially, for SSIM, owing to the input images processed with a kernel 11$\times$11 Gaussian filter, it leads to less 5 pixels near borders around the similarity map. To guarantee image alignment, we exclude each 5 rows and columns for each distorted image border before training SSIM labels.

\subsection{Multi Types Quality Maps Fusion} 

\begin{figure}[h]   
  \centering
  \includegraphics[scale=0.4]{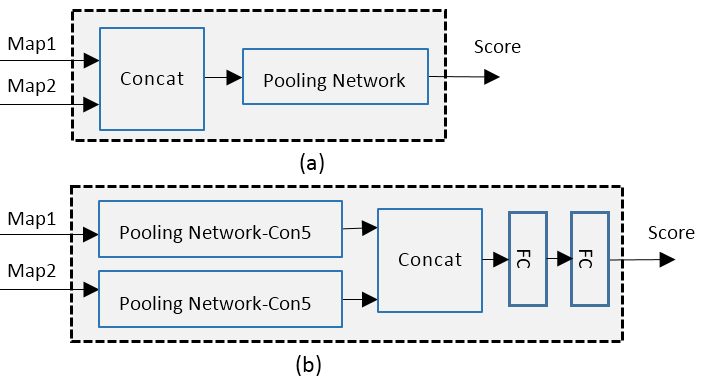}\\
  \caption{Different pooling strategies to combine multi types quality maps: single pooling stream (a), multi pooling streams (b).}\label{fusion}
\end{figure}

For each FR-IQA method, we will train U-Net separately. Many conventional FR-IQA methods~\cite{fsim-15,MDSI1} have demonstrated that multi complementary features are combined to increase the prediction accuracy for image quality. Thus, we also fuse the information from predicted multi types quality maps to feed into the pooling network. Different pooling strategies of quality maps are experimented:

-single pooling stream is performed by concatenating different type quality maps into a multi channels quality map followed by a single pooling network (shown in Fig.~\ref{fusion} (a)).

-multi pooling streams indicate that each type quality map is fed into an independent pooling network. The last convolutional layers of the pooling networks are concatenated, followed by two full connected layers (shown in  Fig. ~\ref{fusion} (b)).

\subsection{Regression}

The input to U-Net is an RGB patch of fixed size 144$\times$144$\times$3 sampled from a distortion image without any image pre-processing. We set the step of the sliding window to 120, i.e. the neighboring patches are overlapped by 24 pixels, which can compensate partial distorted area continuous. Considering that the patch size is large enough to reflect the overall image quality, we set the quality score of each patch to its distorted images subjective ground-truth score. The proposed pooling network is to conduct nonlinear regression from the predicted quality map to the subjective score. To compare the performances of different network structures as regression function, we also test a simple network with only two fully connected layers and ResNet~\cite{resnet}. Then, the final objective function is defined as:
\begin{equation}\label{Ls}
  \centering
  L_{s}(I_{d};\phi,\omega) = ||(f_{\phi}(G_{\omega}(I_{d}))-E\upsilon a)||^{2}
\end{equation}
Where \emph{Eva} denotes the human evaluation for the input distorted image. The final score of a global distorted image is averaging the cropped patches.

\section{Experiments}  

\subsection{Datasets} 

The image quality datasets can be divided into synthetic distortion datasets and authentic distortion datasets according to whether using high-quality reference images. The synthetic distortion dataset is established by generating various distortion types on high-quality images by simulation tools. The images in authentic distortion dataset are not collected by synthetic means, but captured using typical real-world mobile camera devices. Four synthetic distortion datasets including LIVE~\cite{livedata}, CSIQ~\cite{csiq}, TID2008~\cite{tid2008} and TID2013~\cite{tid2013} and three authentic distortion datasets including CLIVE~\cite{Ghadiyaram2015Massive}, CID2013~\cite{Virtanen2015CID2013} and KonIQ-10k~\cite{Lin2018KonIQ} are employed in our experiments to validate the performance of the proposed BPSQM.

The LIVE dataset consists of 982 distorted images with 5 different distortions: white Gaussian noise (WN), Gaussian blur (BLUR), JPEG compression (JPEG), JPEG2000 compression (JP2K) and fast-fading distortion (FF). Each image is associated with Differential Mean Opinion Scores (DMOS) in the range [0, 100]. The CSIQ dataset includes 30 original images and 866 distorted images with 6 distortion types at four to five different levels of distortion. It is reported in the form of DMOS which are normalized to span the range [0, 1]. TID2008 contains 25 reference images and a total of 1700 distorted images with 17 distortion types at 4 degradation levels. TID2013 is an extension of TID2008 and includes seven new types and one more level of distortions. Mean Opinion Scores (MOS) are provided for each image in the range [0, 9]. Owing to more distortion types and images, the TID2013 is more challenging for researchers in the four synthetic distortion datasets.

BPSQM is also evaluated on CID2013, CLIVE and KonIQ-10k. CID2013 ~\cite{Virtanen2015CID2013} containing multiply mixed distortion types has 480 distortion images from 6 specific scenes including different illumination intensities and shooting distances. The dataset is deliberately designed for no reference image quality assessment as all distortion images are camera-captured images in the wild. 188 subjects participated in subjective assessment experiments and MOS values span the range [0,100]. CLIVE~\cite{Ghadiyaram2015Massive} contains 1,162 authentic distortion images with complex mixtures of multiple distortion types. Each image in the dataset is taken by real-world mobile camera devices in natural scenes. An online crowdsourcing system is designed and implemented to collect over 350,000 opinion scores on the images. MOS values lie in the range of [0,100]. To the best of our knowledge, KonIQ-10k is the largest image quality assessment dataset so far, which consists of 10,073 images sampled from around 4.8 million YFCC100m~\cite{Shamma2016YFCC100M} images. Each image has only a MOS value and does not have a corresponding reference image. 1.2 million ratings are collected from 1,467 crowd workers on large scale crowdsourcing experiments to ensure quality scores reliability.

To evaluate the performance of our model, two widely applied correlation criterions are applied in our experiments including the Pearson Linear Correlation Coefficient (PLCC) and Spearman Rank Order Correlation Coefficient (SRCC).For both correlation metrics, a higher value indicates higher performance of a specific quality metric. The MOS values of TID2013 and the DMOS values of CSIQ have been linearly scaled to the same range [0, 100] as the DMOS values in LIVE.

\subsection{Training Method}

The proposed network was implemented in MXNet. U-Net uses a pre-trained model to improve accuracy and speed up convergence. This pre-trained model was trained on a database containing 3,000 high quality images. These images are manually selected from the AVA database~\cite{murray2012ava}. We added four distortion types (Gaussian Blur, JPEG, White Noise and Local Block-Wise Distortion) to these high quality images. Each distortion type has two distortion levels. By outputting the intermediate results, we find that the quality maps from individually training U-Net are more close to the similarity index maps than those from the joint training of the overall framework. Thus, we first only train the generative network on similarity maps, and then fix its parameters in training process of the overall framework.

Our network was trained end-to-end by back-propagation. For optimization, the adaptive moment estimation optimizer (ADAM)~\cite{adam} is employed with $\beta_{1} = 0.9$, $\beta_{2} = 0.999$, $\varepsilon = 10^{-8}$ and $\alpha = 10^{-4}$. We set an initially learning rate to 1$\times$$10^{-3}$ and 5$\times$$10^{-3}$ for the generative network and the pooling network, respectively. We set the weight decay to 1$\times$$10^{-11}$ for all layers to help prevent overfitting. If there is no special emphasis on the following experiments, each dataset is randomly divided into 80\% for training and 20\% for testing by reference images, which ensures that the content of images in test sets never exists in train sets. We only use the horizontally flip operation to expand training data, for general data argumentation skills, such as rotation, zoom, contrast, will affect the final image quality. The models are trained for 100 epochs and we choose the model with the lowest validation error. In order to ensure the fairness of test, we refer to the partitioning strategy of DIQaM~\cite{w-nr-fr} to repeat 10 divisions in the individual dataset test. In the cross-dataset test, the partition was repeated 100 times to eliminate the bias caused by data division, and the detail division strategy is shown in ~\ref{4.J}.

\subsection{Comparison of pixel distortion and patch distortion} 

The core idea of this paper is that pixel-based quality maps correlate better with perceived quality than patch-based quality maps. To compare the performance of the two quality map types, we use the proposed deep pooling network to directly take as input quality maps. The gradient magnitude maps of FSIM generated from the TID2013 dataset and corresponding variant patch maps are selected as the training data. Eight different local patch sizes (1, 2, 4, 8, 16, 24, 36, 48)  are applied in the experiment. Each pixel in patch-based quality map is equal to the average value of a local patch. In particular, the local patch size 1 represents the pixel-based quality map.

As shown in  Table~\ref{table1}, the performance decreases gradually as the local patch size increases. The pixel-based quality maps (size=1) get results SRCC=0.928 PLCC=0.934, which is better than other patch-based quality maps. If we take average operations to disturb the relationship between neighboring pixels, it would degrade the final quality assessment performance. This gives a powerful proof that each distortion pixel value has a close correlation with the perceived quality. Therefore, how to accurately predict a pixel-based quality map is the key idea of this work.

\begin{table}[t]
  \centering
  \caption{Srocc and PLCC comparison for different sizes of average local patches. ALL models are trained on the gradient magnitude map of FSIM generated from the TID2013 Dataset}\label{table1}
  \resizebox{\linewidth}{!}{
  \begin{tabular}{c|cccccccc}
  \hline
  \hline
         &    1   &	2	&	4	&	8	& 	16	&	24	&	36	&	48\\ \hline
   SRCC  & 0.928  & 0.916  & 0.900   & 0.890    & 0.870 	&	0.856	&	0.837	&	0.826\\
   PLCC  & 0.934  & 0.924  & 0.911   & 0.898    & 0.873 	&	0.861	&	0.861	&	0.830 \\ \hline\hline
  \end{tabular}}
\end{table}

\subsection{Quality Map Prediction} 

\begin{figure*}[pt]  
  \centering
\subfigure{
\begin{minipage}{4cm}
\centerline{\includegraphics[width=3.5cm]{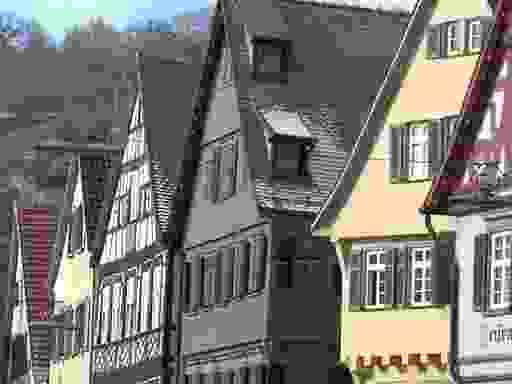}}\vspace{0.25cm}
\centerline{\scriptsize{(A)}}
\end{minipage}}
\subfigure{
\begin{minipage}{4cm}
\centerline{\includegraphics[width=3.5cm]{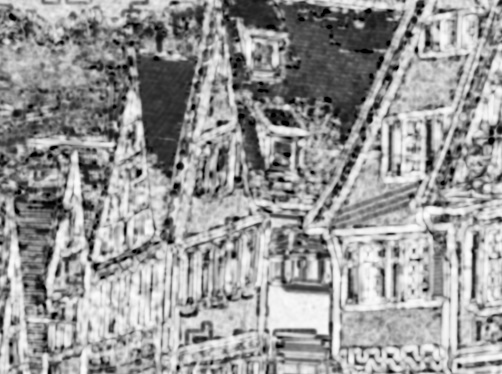}}\vspace{0.25cm}
\centerline{\scriptsize{A(1)}}
\vspace{0.25cm}
\centerline{\includegraphics[width=3.50cm]{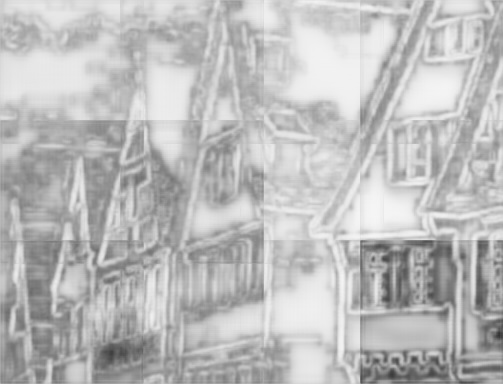}}\vspace{0.25cm}
\centerline{\scriptsize{A(4)}}
\vspace{0.25cm}
\end{minipage}}
\subfigure{
\begin{minipage}{4cm}
\centerline{\includegraphics[width=3.50cm]{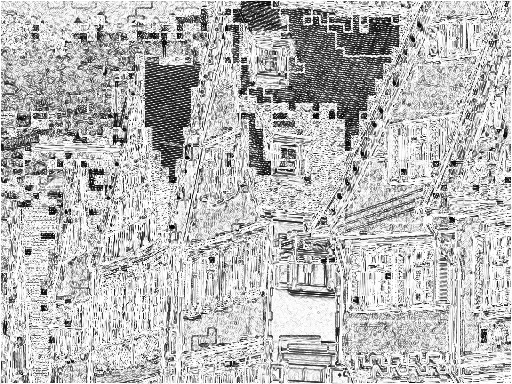}}\vspace{0.25cm}
\centerline{\scriptsize{A(2)}}
\vspace{0.25cm}
\centerline{\includegraphics[width=3.50cm]{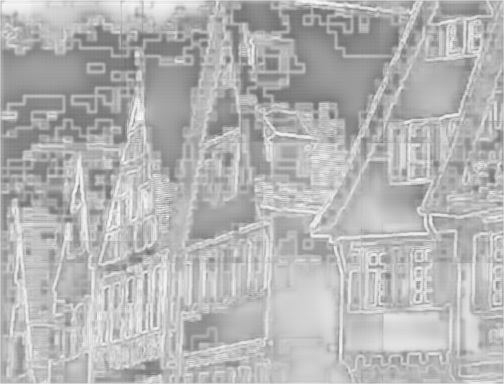}}\vspace{0.25cm}
\centerline{\scriptsize{A(5)}}
\vspace{0.25cm}
\end{minipage}}
\subfigure{
\begin{minipage}{4cm}
\centerline{\includegraphics[width=3.50cm]{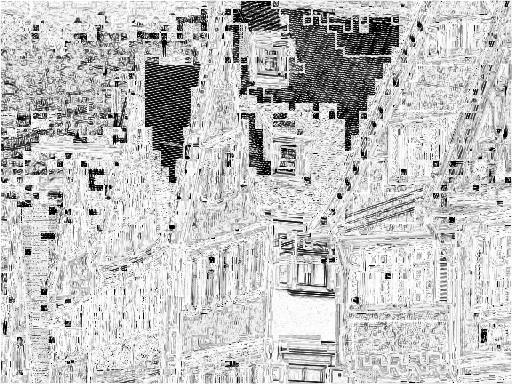}}\vspace{0.25cm}
\centerline{\scriptsize{A(3)}}
\vspace{0.25cm}
\centerline{\includegraphics[width=3.50cm]{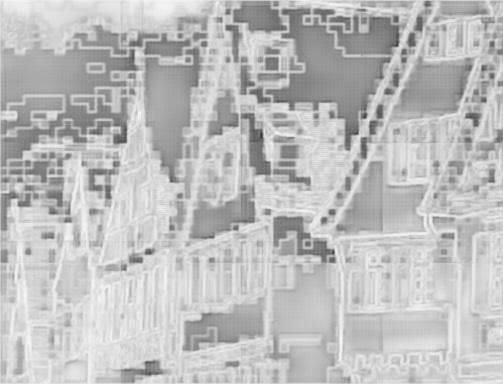}}\vspace{0.25cm}
\centerline{\scriptsize{A(6)}}
\vspace{0.25cm}
\end{minipage}}

\subfigure{
\begin{minipage}{4.0cm}\vspace{-1.0em}
\centerline{\includegraphics[width=3.50cm]{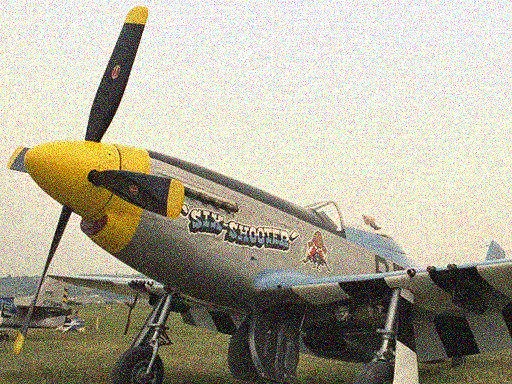}}\vspace{0.25cm}
\centerline{\scriptsize{(B)}}
\end{minipage}}
\subfigure{
\begin{minipage}{4.0cm}\vspace{-1.0em}
\centerline{\includegraphics[width=3.50cm]{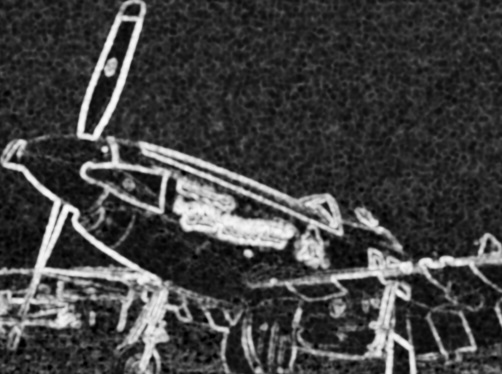}}\vspace{0.25cm}
\centerline{\scriptsize{B(1)}}
\vspace{0.25cm}
\centerline{\includegraphics[width=3.50cm]{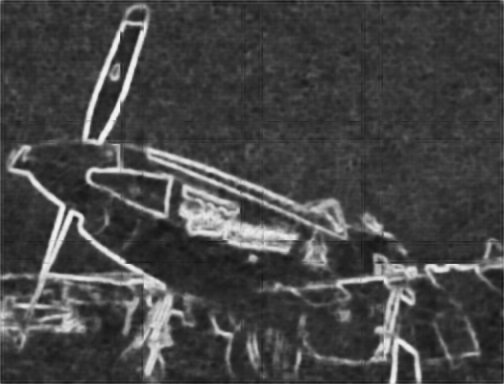}}\vspace{0.25cm}
\centerline{\scriptsize{B(4)}}
\vspace{0.25cm}
\end{minipage}}
\subfigure{
\begin{minipage}{4.0cm}\vspace{-1.0em}
\centerline{\includegraphics[width=3.50cm]{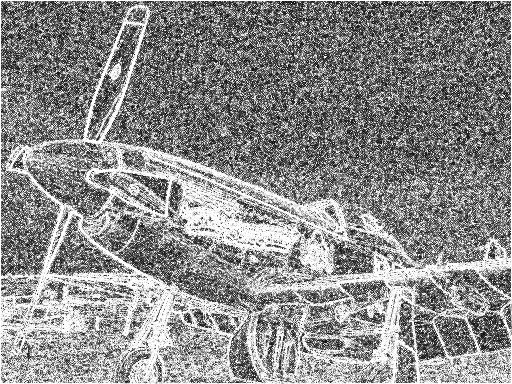}}\vspace{0.25cm}
\centerline{\scriptsize{B(2)}}
\vspace{0.25cm}
\centerline{\includegraphics[width=3.50cm]{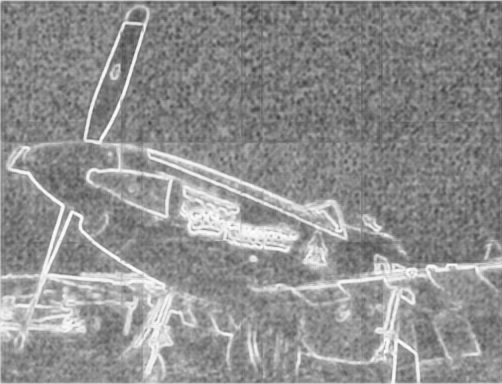}}\vspace{0.25cm}
\centerline{\scriptsize{B(5)}}
\vspace{0.25cm}
\end{minipage}}
\subfigure{
\begin{minipage}{4.0cm}\vspace{-1.0em}
\centerline{\includegraphics[width=3.50cm]{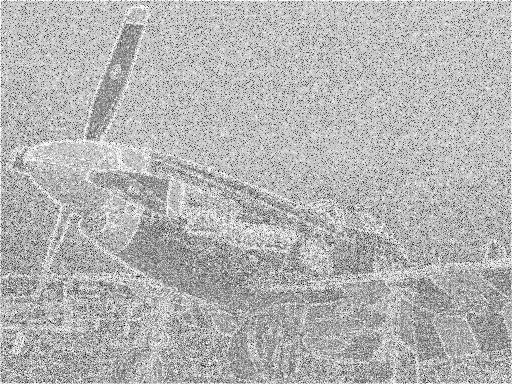}}\vspace{0.25cm}
\centerline{\scriptsize{B(3)}}
\vspace{0.25cm}
\centerline{\includegraphics[width=3.50cm]{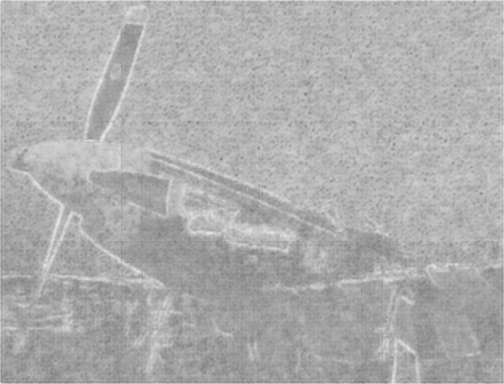}}\vspace{0.25cm}
\centerline{\scriptsize{B(6)}}
\vspace{0.25cm}
\end{minipage}}

\subfigure{
\begin{minipage}{4.0cm}\vspace{-1.0em}
\centerline{\includegraphics[width=3.50cm]{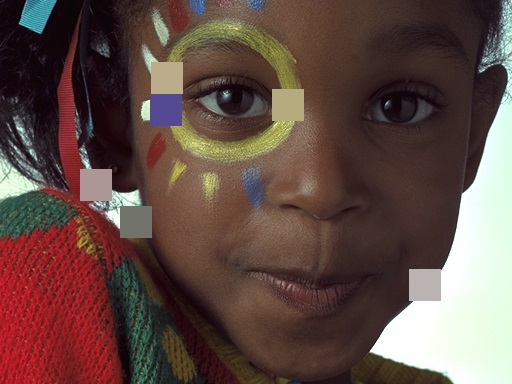}}\vspace{0.25cm}
\centerline{\scriptsize{(C)}}
\end{minipage}}
\subfigure{
\begin{minipage}{4.0cm}\vspace{-1.0em}
\centerline{\includegraphics[width=3.50cm]{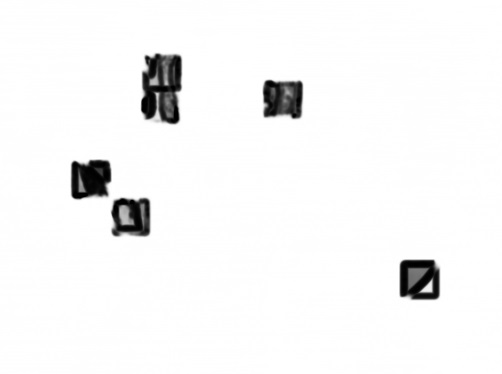}}\vspace{0.25cm}
\centerline{\scriptsize{C(1)}}
\vspace{0.25cm}
\centerline{\includegraphics[width=3.50cm]{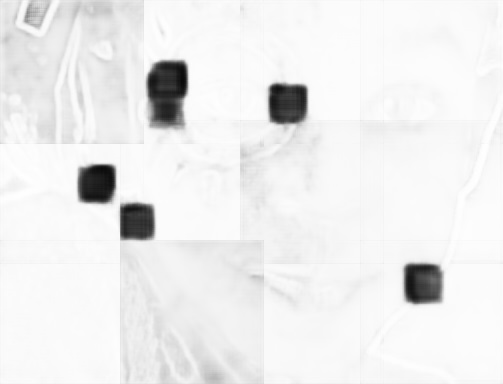}}\vspace{0.25em}
\centerline{\scriptsize{C(4)}}
\vspace{0.25cm}
\end{minipage}}
\subfigure{
\begin{minipage}{4.0cm}\vspace{-1.0em}
\centerline{\includegraphics[width=3.50cm]{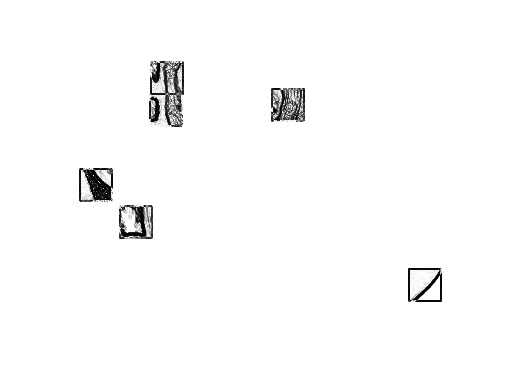}}\vspace{0.25em}
\centerline{\scriptsize{C(2)}}
\vspace{0.25cm}
\centerline{\includegraphics[width=3.50cm]{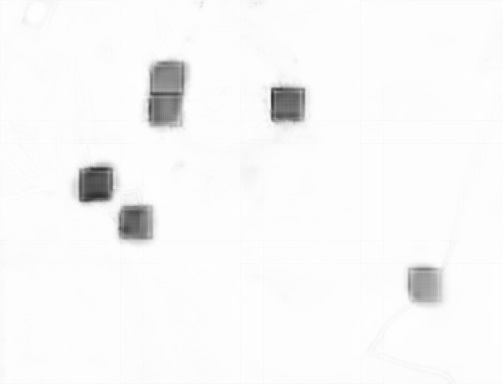}}\vspace{0.25em}
\centerline{\scriptsize{C(5)}}
\vspace{0.25cm}
\end{minipage}}
\subfigure{
\begin{minipage}{4.0cm}\vspace{-1.0em}
\centerline{\includegraphics[width=3.50cm]{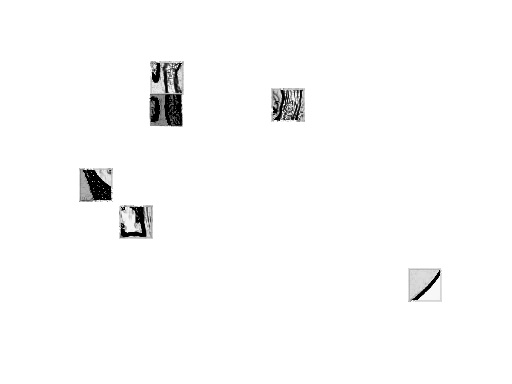}}\vspace{0.25em}
\centerline{\scriptsize{C(3)}}
\vspace{0.25cm}
\centerline{\includegraphics[width=3.50cm]{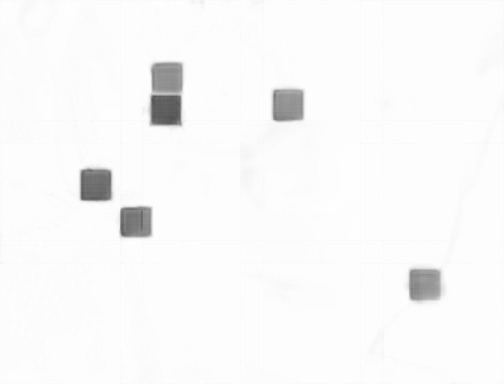}}\vspace{0.25em}
\centerline{\scriptsize{C(6)}}
\vspace{0.25cm}
\end{minipage}}
\caption{Predicted quality maps and ground-truth similarity maps: (A), (B) and (C) are distorted images with JPEG, HFN, LBWD, respectively; the second, the third and the forth columns indicate three FR-IQA methods which are SSIM, Fg and MDSI, respectively. A(1-3), B(1-3) and C(1-3) are ground-truth similarity maps. A(4-6), B(4-6) and C(4-6) are predicted quality maps.}\label{fig:fsim}
\end{figure*}

\begin{figure*}[h]  
\begin{center}
\subfigure[0.917]{
\begin{minipage}{3.0cm}
\includegraphics[width=3.0cm]{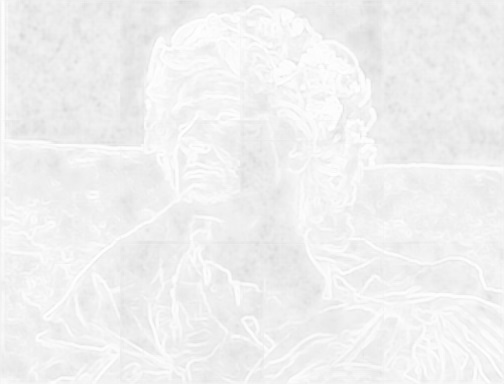}\vspace{0.25cm}
\end{minipage}}
\subfigure[0.893]{
\begin{minipage}{3cm}
\centering
\includegraphics[width=3.0cm]{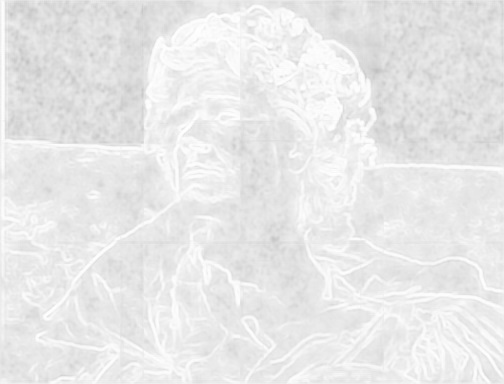}\vspace{0.25cm}
\end{minipage}}
\subfigure[0.847]{
\begin{minipage}{3.0cm}
\centering
\includegraphics[width=3.0cm]{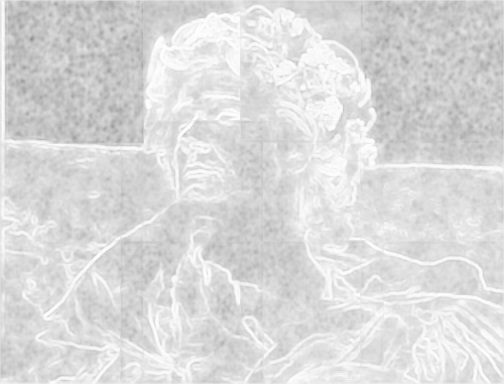}\vspace{0.25cm}
\end{minipage}}
\subfigure[0.775]{
\begin{minipage}{3.0cm}
\centering
\includegraphics[width=3.0cm]{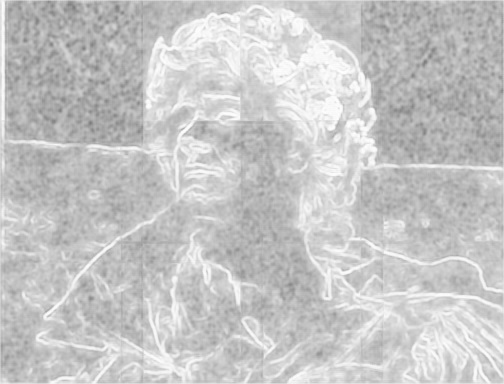}\vspace{0.25cm}
\end{minipage}}
\subfigure[0.670]{
\begin{minipage}{3.0cm}
\centering
\includegraphics[width=3.0cm]{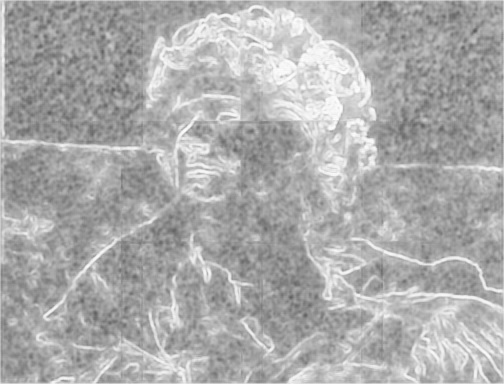}\vspace{0.25cm}
\end{minipage}}
\subfigure[0.931]{
\begin{minipage}{3.0cm}
\includegraphics[width=3.0cm]{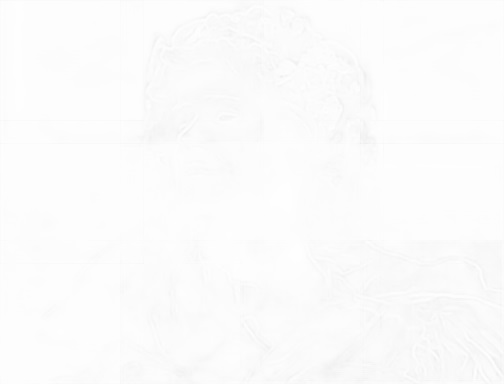}\vspace{0.25cm}
\end{minipage}}
\subfigure[0.917]{
\begin{minipage}{3.0cm}
\centering
\includegraphics[width=3.0cm]{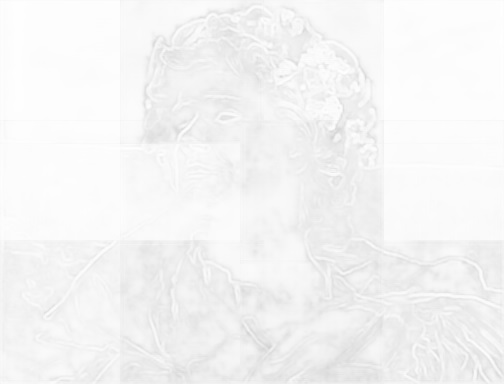}\vspace{0.25cm}
\end{minipage}}
\subfigure[0.882]{
\begin{minipage}{3.0cm}
\centering
\includegraphics[width=3.0cm]{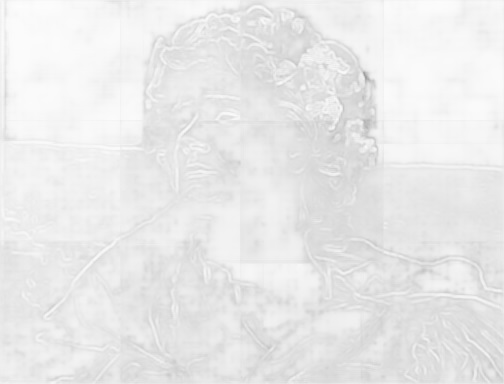}\vspace{0.25cm}
\end{minipage}}
\subfigure[0.819]{
\begin{minipage}{3.0cm}
\centering
\includegraphics[width=3.0cm]{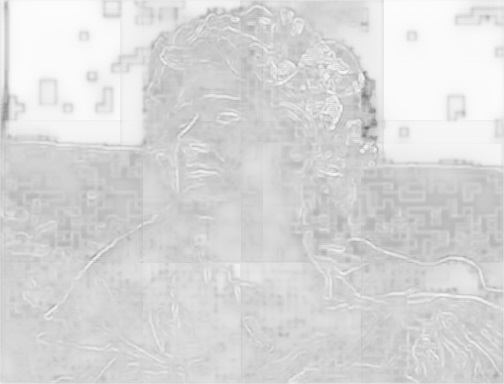}\vspace{0.25cm}
\end{minipage}}
\subfigure[0.737]{
\begin{minipage}{3.0cm}
\centering
\includegraphics[width=3.0cm]{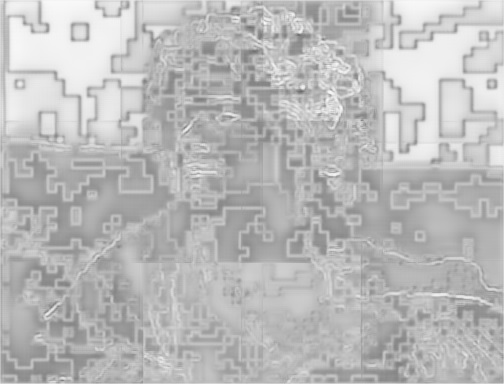}\vspace{0.25cm}
\end{minipage}}
\caption{Examples of predicted quality maps with various distortion levels of spatially correlated noise, and JPEG: (a)-(e) are distorted by spatially correlated noise; (f)-(j) are distorted by JPEG. The values indicate the predicted scores output from the pooling network. Smaller values indicate higher distortions.}\vspace{-0.3cm}
\label{fig:res}
\end{center}
\end{figure*}

To validate if BPSQM is consistent with human visual perception, the intermediate generative quality maps and their corresponding similarity map labels are shown in Fig. ~\ref{fig:fsim}. The first column in Fig. ~\ref{fig:fsim} shows three different distortion types from TID2013, including JPEG, high frequency noise (HFN) and local block-wise distortions (LBWD). The remaining columns correspond to SSIM, the gradient magnitude of FSIM (Fg) and MDSI, respectively. A(1-3), B(1-3) and C(1-3) are the ground-truth map labels. A(4-6), B(4-6) and C(4-6) are the predicted quality maps. The dark areas indicate distorted pixels. Overall, the generative quality maps are similar with ground-truth maps on distorted degrees and areas. In case of JPEG distortion, the artifact edges caused by compression on the root are clearly shown in A(5) and A(6). But owing to SSIM similarity index maps emphasizing local structure features, this leads to some areas in the predicted quality map to be smooth without local pixels distortion, as shown in A(4). For HFN, noises spread over an overall distorted image. B(4-6) not only display the uniform distribution well but also give a clear airplane profile. LBWD is a very challenging distortion type for most BIQA methods to distinguish additional blocks and undistorted regions. Even though some wrong pixel distortion predictions appear in the undistorted regions, as shown in C(4-6), each predicted block is darker than other areas. Meanwhile, the predicted positions of local blocks can agree with those of the map labels.

In Fig. ~\ref{fig:res}, the predicted quality maps of spatially correlated noise and JPEG with different distortion levels are shown. The first row denotes the spatially correlated noise, and the second row denotes the JPEG. With the noises becoming strong gradually from left to right, the predicted quality maps grow darker and darker as shown in (a)-(e). Meanwhile, when the degree of JPEG compression increases, the blocking artifact on the sculpture area was emphasized in (j). Generally, with the degree of distortion increasing, the predicted scores gradually decrease, which suggests that BPSQM predicts good pixel-by-pixel quality maps agreeing with the distortion correlations between the reference and distorted images.

\subsection{Dependency on FR-IQA Similarity Map}  

To validate the feasibility and effectiveness of directly pooling quality maps, we compare pooling ground-truth FR-IQA maps with pooling predicted quality maps. We choose the more challenging full TID2013 dataset in this experiment. Here, directly pooling ground-truth map labels from SSIM and the gradient magnitude of FSIM are denoted by S\_LB and Fg\_LB, respectively. Directly pooling predicted maps of SSIM is denoted by S\_PM. The gradient magnitude computed from FSIM is denoted by Fg\_PM. We also feed the distorted image into the pooling network directly, named as D\_LB. The results are shown in Table~\ref{table2}, we can see that the S\_LB and Fg\_LB both perform better than their original methods, especially for the SRCC of SSIM increasing from 0.637 to 0.904, which suggests that the deep pooling network can better fit a quality map to a subjective score than simple averaging. The D\_LB performs worse than Fg\_PM. We consider the primary reason is that distorted images contain too much redundant information and do not highlight distorted distribution features. Even though the deep network has strong ability of extracting discriminative features, it is still not enough to accurately present distorted patterns. For this reason, we need to firstly predict similar quality maps which correctly reveal the distorted areas and degrees.

\begin{table} [t]
  \centering
  \caption{SRCC and PLCC comparison for pooling ground-truth FR-IQA maps and pooling predicted quality maps}\label{table2}  
  \resizebox{\linewidth}{!}{
  \begin{tabular}{c|ccc|cccc}
  \hline
  \hline
        & \multicolumn{3}{c|}{NR} & \multicolumn{4}{c}{FR} \\ \hline 
        & D\_LB & S\_PM & Fg\_PM & S\_LB & Fg\_LB & SSIM  & FSIMc\\ \hline
  SRCC  & 0.781 & 0.758 & 0.828     & 0.904 & 0.923 & 0.637 & 0.851\\
  PLCC  & 0.837 & 0.803 & 0.856     & 0.913 & 0.930 & 0.691 & 0.877\\ \hline
  \hline
  \end{tabular}}
\end{table}

\begin{table}[t] 
  \centering
  \caption{SRCC and PLCC comparison for each predicted quality map from different FR-IQA methods on the TID2013 dataset}\label{table3}
  \begin{tabular}{c|ccccc}
  \hline
  \hline
         & No\_LB & S\_PM & Fg\_PM & Fp\_PM & MD\_PM \\ \hline
   SRCC  & 0.736  & 0.758 & 0.828  & 0.723  & 0.863 \\
   PLCC  & 0.779  & 0.803 & 0.856  & 0.789  & 0.879 \\ \hline
   \hline
  \end{tabular}\vspace{-0.3cm}
\end{table}

\subsection{Similarity Map Labels Comparison} 

\begin{figure}[t]
\begin{center}
\subfigure[]{
  \begin{minipage}{4cm}
  \includegraphics[width=4.0cm,height=2.8cm]{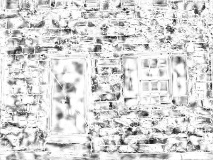}\vspace{0.25cm}\label{Fig1:a}
\end{minipage}}
\subfigure[]{
  \begin{minipage}{4cm}
  \includegraphics[width=4.0cm,height=2.8cm]{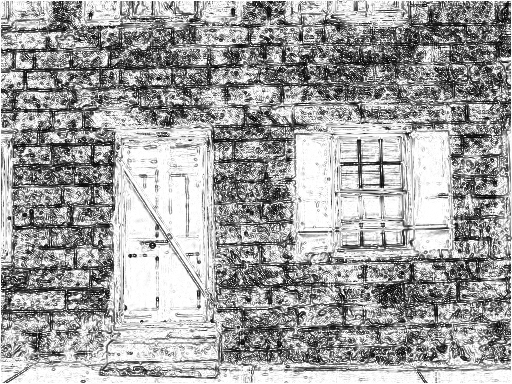}\vspace{0.25cm}\label{Fig1:b}
\end{minipage}}
 \caption{(a) is a quality map of phase congruency features from FSIM. (b) is a quality map of gradient features from FSIM. The two images reflect a same distorted image.}\label{fig:WALL} 
\vspace{-0.75cm}
\end{center}
\end{figure}

To investigate the performance of different FR-IQA ground-truth maps, the similarity maps derived from SSIM, FSIM and MDSI were respectively chosen as labels for training the model. The TID2013 dataset with all distortion types was applied in this experiment. The combination of gradient and chromaticity similarity maps from MDSI are referred to MD\_PM. The phase congruency (PC) from FSIM is denoted by Fp\_PM. In order to analyze the effect of removing FR-IQA similarity maps, we directly employ the overall framework to perform an end-to-end training without any FR-IQA map labels, referred to No\_LB. The final SRCC and PLCC values are shown in Table~\ref{table3}, No\_LB achieves worse performance among the results except for Fp\_PM. Clearly, selecting FR-IQA similarity maps for training the generative network can help to learn a better model for predicting image quality, because the task improves the ability of U-Net learning discriminative features about distortions. In particular, Fp\_PM achieves unfavorable performance and seems not to fit with this framework. In FSIM, the PC is used both as features and a weighting function. But when it comes to the the ability in describing local pixel distortion, the selected quality map of PC from FSIM is less than the quality map of gradient features from FSIM. Fig. ~\ref{fig:WALL} shows the differences between the two quality map types. In contrast, Fg\_PM achieves the second rank, suggesting that gradient distortion variations learned from the U-Net is more suitable than phase congruency to apply into the framework. MD\_PM performs the best result, suggesting that the chromaticity feature can be complementary to gradient features.

\subsection{Effects of Pooling Network}  

To investigate the pooling performance of convolutional network with different depth and full connected layers. We experimented with three pooling network structures. The first network is the pooling network proposed by this paper, called DPN. The second network only contains two full connected layers with 1024 neurons each, called FC2. ResNets~\cite{resnet} with 18 layers and 50 layers are selected as the third and the forth pooling network. The results are shown in Table~\ref{table4}. The DPN performs better than FC2, which indicates that the additional convolutional layers have the strong ability of pooling quality maps. Although ResNet performs better than shallow networks on image classification and recognition, the deeper network is no necessarily to use in this experiment.
This well indicates that the proposed framework not mainly depends on the depth layer of convolutional network, but the focus is on the quality of feature quality maps predicted by generative network.

\subsection{Quality Maps Fusion}  

We evaluate two different fusion schemes for combining multi types predicted quality maps. The first is a single pooling stream scheme (shown in  Fig. ~\ref{fusion} (a)) and the second is multi pooling streams (shown in Fig. ~\ref{fusion} (b)). The results are given in Table~\ref{table5}. We can see that the single pooling stream is better. This further illustrates that, compared with the feature maps output from the pooling network-con5, the quality maps directly output from U-Net are more in consistent with human vision.

\begin{table}[h] 
  \centering
  \caption{SRCC and PLCC comparison for different pooling networks using Fg\_PM on the TID2013 dataset}\label{table4}
  \begin{tabular}{c|cccc}
  \hline \hline
       & DPN   & FC2   & ResNet18 & ResNet50 \\ \hline
  SRCC & 0.828 & 0.707 & 0.787    & 0.795    \\
  PLCC & 0.856 & 0.711 & 0.833    & 0.840    \\
  \hline \hline
  \end{tabular}
\end{table}
\begin{table}[h] 
  \centering
  \caption{SRCC and PLCC comparison for different fusion schemes and multi predicted quality maps combinations on TID2013}\label{table5}
  \begin{tabular}{c|c|cccc}
  \hline \hline
  & & Fg\_MD & S\_Fg & MD\_S & S\_Fg\_MD \\ \hline
  \multirow{2}{*}{Single stream} & SRCC & 0.862 & 0.825 & 0.853 & 0.855 \\
  & PLCC & 0.885 & 0.859 & 0.873 & 0.880 \\ \hline
  \multirow{2}{*}{Multi streams} & SRCC & 0.842 & 0.821 & 0.825 & 0.834 \\
  & PLCC & 0.873 & 0.854 & 0.861 & 0.868 \\ \hline \hline
  \end{tabular}
\end{table}

In the experiment \emph{F} section, S\_PM, MD\_PM and Fg\_PM rank the top three. In order to compare multi type quality maps combinations, we design all possible combinations which contain any two types and all types. As we can see from Table~\ref{table5}, the combination of more types seems not yield significant performance gains. Moreover, the predicted SSIM quality map combined with other quality maps caused a little performance degradation. Since the gradient operator is different in MDSI and FSIM, the two joint achieves the best performance, which indicates that gradient features are complementary to each other.
\begin{table*}[t]  
  \centering
  \caption{Performance comparison on the whole dataset (LIVE, CSIQ and TID2013). Italics indicate our proposed model.}\label{table6}
  \resizebox{\linewidth}{!}{
\begin{tabular}{cc|ccc|cccccccccc}
\hline\hline
Type   &  & \multicolumn{3}{c|}{FR} & \multicolumn{10}{c}{NR} \\ \hline
Method &  &  SSIM & FSIMc &MDSI & DIIVINE & BRISQUE & NIQE & IMNSS & HFD-BIQA & BIECON & DIQaM & \emph{BPSQM-Fg} & \emph{BPSQM-MD} & \emph{BPSQM-Fg-MD} \\ \hline
\multirow{2}{*}{LIVE IQA} & SRCC  & 0.948 & 0.960& 0.966 & 0.892 & 0.929 & 0.908 & 0.943 & 0.951 & 0.958 & 0.960 & 0.971 & 0.967 & \textbf{0.973} \\
 & PLCC  & 0.945 & 0.961 & 0.965& 0.882 & 0.920 & 0.908 & 0.944 & 0.948 & 0.960 & \textbf{0.972} & 0.961 & 0.955 & 0.963 \\\hline
\multirow{2}{*}{  CSIQ  } & SRCC  & 0.876 & 0.931& 0.956 & 0.804 & 0.812 & 0.812 & 0.825 & 0.842 & 0.815 & - & 0.862 & 0.860 & \textbf{0.874} \\
 & PLCC  & 0.861 & 0.919 & 0.953& 0.776 & 0.748 & 0.629 & 0.789 & 0.890 & 0.823 & - & 0.891 & 0.904 & \textbf{0.915} \\\hline
\multirow{2}{*}{TID2013}  & SRCC  & 0.637 & 0.851 & 0.889& 0.643 & 0.626 & 0.421 & 0.598 & 0.764 & 0.717 & 0.835 & 0.828 & \textbf{0.863} & 0.862 \\
 & PLCC  & 0.691 & 0.877& 0.908 & 0.567 & 0.571 & 0.330 & 0.522 & 0.681 & 0.762 & 0.855 & 0.856 & 0.879 & \textbf{0.885} \\\hline
\multirow{2}{*}{Weighted Avg.}  & SRCC  & 0.743 & 0.887& 0.917 & 0.722 & 0.721 & 0.589 & 0.708 & 0.816 & 0.783 & - & 0.863 & 0.884 & \textbf{0.887} \\
 & PLCC  & 0.773 & 0.902 & 0.928& 0.668 & 0.673 & 0.500 & 0.655 & 0.772 & 0.813 & - & 0.884 & 0.899 & \textbf{0.906} \\\hline
 \hline \hline
\end{tabular}}
\end{table*}

\begin{table*} 
 \caption{SRCC comparison on individual distortion types across 10 sessions on the LIVE IQA and TID2008 datasets. Italics indicate deep learning-based methods.}\label{table7}
  \resizebox{\linewidth}{!}{
  \begin{tabular}{c|ccccc|ccccccccccccc}
  \hline \hline
  \multirow{2}{*}{Method} & \multicolumn{5}{c|}{LIVE IQA} & \multicolumn{13}{c}{TID2008} \\ \cline{2-19}
   & JP2K & JPEG & WN & BLUR & FF & AGN & ANMC & SCN & MN & HFN & IMN & QN & GB & DEN & JPEG & JP2K & JGTE & J2TE \\ \hline
SSIM & 0.961 & 0.972 & 0.969 & 0.952 & \textbf{0.956}  & 0.811 & 0.803 & 0.792 & 0.852 & 0.875 & 0.700 & 0.807 & 0.903 & 0.938 & 0.936 & 0.906 & 0.840 & 0.800 \\
GMSD & 0.968 & 0.973 & \textbf{0.974} & 0.957 & 0.942 & \textbf{0.911} & \textbf{0.878} & \textbf{0.914} & 0.747 & 0.919 & 0.683 & 0.857 & 0.911 & \textbf{0.966} & \textbf{0.954} & \textbf{0.983} & 0.852 & 0.873 \\
FSIMc & \textbf{0.972} & \textbf{0.979} & 0.971 & \textbf{0.968} & 0.950 & 0.910 & 0.864 & 0.890 & \textbf{0.863} & \textbf{0.921} & \textbf{0.736} & \textbf{0.865} & \textbf{0.949} & 0.964 & 0.945 & 0.977 & \textbf{0.878} & \textbf{0.884} \\ \hline
   BLIINDSII & 0.929 & 0.942 & 0.969 & 0.923 & 0.889 & 0.779 & 0.807 & 0.887 & 0.691 & 0.917 & 0.908 & 0.851 & 0.952 & 0.908 & 0.928 & 0.940 & 0.865 & \textbf{0.855} \\
   DIIVINE & 0.937 & 0.910 & 0.984 & 0.921 & 0.863 & 0.812 & 0.844 & 0.854 & 0.713 & 0.922 & 0.915 & 0.874 & 0.943 & 0.912 & 0.930 & 0.938 & 0.873 & 0.852 \\
   BRISQUE & 0.914 & 0.965 & 0.979 & 0.951 & 0.877 & 0.853 & 0.861 & 0.885 & 0.810 & 0.931 & 0.927 & 0.881 & 0.933 & \textbf{0.924} & 0.934 & 0.944 & \textbf{0.891} & 0.836 \\
   NIQE    & 0.914 & 0.937 & 0.967 & 0.931 & 0.861 & 0.786 & 0.832 & 0.903 & 0.835 & 0.931 & 0.913 & 0.893 & \textbf{0.953} & 0.917 & 0.943 & 0.956 & 0.862 & 0.827 \\
   \emph{BIECON} & 0.952 & \textbf{0.974} & 0.980 & 0.956 & 0.923 & 0.913 & 0.835 & 0.903 & 0.835 & 0.931 & 0.913 & 0.893 & \textbf{0.953} & 0.917 & 0.943 & 0.956 & 0.862 & 0.827 \\
   \emph{BPSQM-Fg-MD} & 0.969 & 0.946 & \textbf{0.993} & \textbf{0.986} & 0.960 & 0.881 & 0.801 & 0.935 & 0.786 & 0.938 & \textbf{0.933} & \textbf{0.920} & 0.937 & 0.914 & 0.943 & \textbf{0.967} & 0.829 & 0.644 \\
   \emph{BPSQM-MD}   & \textbf{0.972} & 0.929 & 0.985 & 0.977 & \textbf{0.964} & \textbf{0.923} & \textbf{0.880} & \textbf{0.941} & \textbf{0.948} & \textbf{0.948} & 0.892 & 0.909 & 0.908 & 0.878 & \textbf{0.950} & \textbf{0.967} & 0.836 & 0.756 \\
   \hline \hline
  \end{tabular}}
\end{table*}

\subsection{Performance Comparison} 

In Table~\ref{table6}, the proposed BPSQM is compared with 7 state-of-the-art NR-IQA methods (DIIVINE~\cite{DIIVINE-18}, BRISQUE~\cite{no--3}, NIQE~\cite{NIQE-21}, IMNSS~\cite{IMNSS}, HFD-BIQA~\cite{iccv2017}, BIECON~\cite{kimfully} and DIQaM~\cite{w-nr-fr}) and 3 FR-IQA methods (MDSI~\cite{MDSI1}, SSIM~\cite{4--1}, FSIM~\cite{fsim-15}). All distortion types are considered over the three datasets. The best PLCC and SRCC for the NR IQA methods are highlighted. The weighted average in the last column is proportional to the number of distorted images of each dataset. We can see that the BPSQM obtains superior performance to state-of-the-art BIQA methods, except for DIQaM evaluated by PLCC on LIVE. Especially for the challenging TID2013, BPSQM achieves a remarkable improvement against BEICON. It is obvious that predicted global quality maps in pixel level helps the model extract more useful features to achieve a good accuracy. Meanwhile, the BPSQM-Fg-MD achieves competitive performance to some FR-IQA methods.

Table~\ref{table7} shows the SRCC performance of the competing BIQA methods for distortion types on LIVE and TID2008 dataset. The best results are in bold. In general, BPSQM-MD and BPSQM-Fg-MD achieve the competitive performances among most distortion types on the two datasets. Compared with BIECON, BPSQM is more capable in dealing with the distortion of SCN, HFN, QN and JP2K. By contrast, for the distortions of GB, JGTE and J2TE, BEICON performs better than BPSQM-MD and BPSQM-Fg-MD. Moreover, the achieved scores on some distortion types are close to state-of-the-art FR-IQA methods.

\begin{figure*}[pt] 
\centering
\subfigure{
\begin{minipage}{4cm}
\centerline{\includegraphics[width=3.50cm]{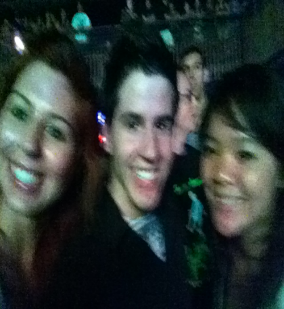}}\vspace{0.25cm}
\centerline{\scriptsize{A(1)}}
\vspace{0.25cm}
\centerline{\includegraphics[width=3.50cm]{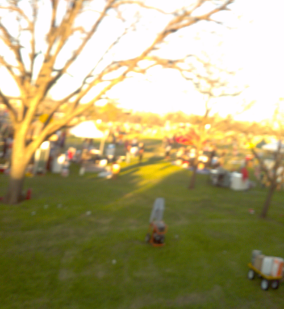}}\vspace{0.25cm}
\centerline{\scriptsize{C(1)}}
\vspace{0.25cm}
\end{minipage}}
\subfigure{
\begin{minipage}{4cm}
\centerline{\includegraphics[width=3.50cm]{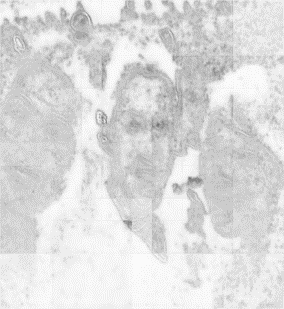}}\vspace{0.25cm}
\centerline{\scriptsize{A(2)}}
\vspace{0.25cm}
\centerline{\includegraphics[width=3.50cm]{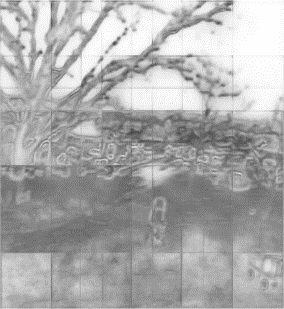}}\vspace{0.25cm}
\centerline{\scriptsize{C(2)}}
\vspace{0.25cm}
\end{minipage}}
\subfigure{
\begin{minipage}{4cm}
\centerline{\includegraphics[width=3.50cm]{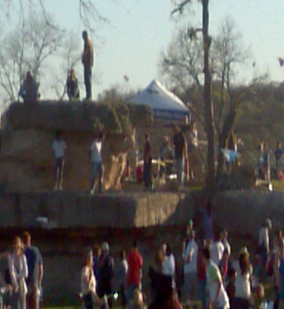}}\vspace{0.25cm}
\centerline{\scriptsize{B(1)}}
\vspace{0.25cm}
\centerline{\includegraphics[width=3.50cm]{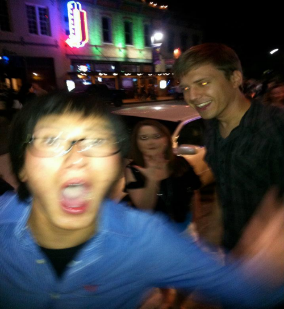}}\vspace{0.25cm}
\centerline{\scriptsize{D(1)}}
\vspace{0.25cm}
\end{minipage}}
\subfigure{
\begin{minipage}{4cm}
\centerline{\includegraphics[width=3.50cm]{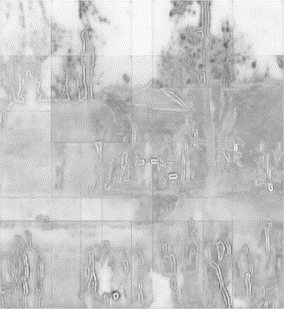}}\vspace{0.25cm}
\centerline{\scriptsize{B(2)}}
\vspace{0.25cm}
\centerline{\includegraphics[width=3.50cm]{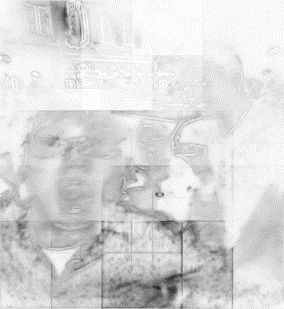}}\vspace{0.25cm}
\centerline{\scriptsize{D(2)}}
\vspace{0.25cm}
\end{minipage}}
\caption{Distortion images and predicted quality maps : A(1), B(1), C(1) and D(1) are authentic distortion images in CLIVE dataset; A(2), B(2), C(2) and D(2) are predicted quality maps.}\label{fig:nature}
\end{figure*}

\subsection{Performance on authentic datasets} 

In order to test the performance of BPSQM on authentic distortion datasets, we perform end-to-end training from image to score on the CLIVE, CID2013 and Koniq10k. The training methods are divided into two forms: we select the pre-trained generative network model on TID2013 with FSIM features as initial parameters, named as BPSQM-Pre. The other is that the entire network chooses Xavier as the initialization strategy, named as BPSQM-Int. For a fair and reasonable comparison, according to the splitting strategy of deepIQA~\cite{w-nr-fr}, we divided CLIVE into 698 for training, 232 for verification, and 232 for testing. Meanwhile, according to the CIQM~\cite{Tang2017An} division strategy, CID2013 is randomly divided into 80\% for training and 20\% for testing. The median value of ten random partition results is used as the final value.

\begin{table}[t]  
  \centering
  \caption{The SRCC,PLCC and RMSE results of the proposed CIQM model 
and the compared metrics on authentically distorted image datasets. 
The top two metrics are highlighted in boldface.}
\label{table9}
  \resizebox{\linewidth}{!}{
\begin{tabular}{c|cc|cc|cc}
\hline\hline
Dataset		 & \multicolumn{2}{c}{CLIVE} & \multicolumn{2}{c}{CID2013} & \multicolumn{2}{c}{Koniq10k}\\ \hline
IQM-Name	      &      PLCC &  SRCC        &      PLCC &  SRCC         &    PLCC &  SRCC        \\ \hline
BRISQUE	     	& 0.610&0.602		&	0.751&0.750		&	0.704&0.700		\\
DIIVINE		& 0.557&0.509		&	0.608&0.578		&	0.622&0.585		\\
BLINDS-II    	& 0.449&0.404		&	0.715&0.701		&	0.583&0.575		\\
FRIQUEE-all	& 0.720& \textbf{0.720}		&	- 	& - 			&	- 	& - 			\\    
Kang-2014		& \textbf{0.730}&0.710		&	- 	& - 			&	0.670&0.630		\\
SSEQ~\cite{Liu2014No}     &	0.539&0.496	&	0.703&0.690		&	0.615&0.586		\\
BIQI~\cite{Moorthy2010A}  & 	0.532&0.507	& 	0.774&0.746		& 	0.619&0.545		\\
CIQM~\cite{Tang2017An}    & 	0.618&0.583	&	0.817&0.804		&	- 	& - 			\\ 
WaDIQaM-NR		& 0.680&0.671		&	- 	& - 			&	- 	& - 			\\ \hline
BPSQM-Int	& 0.707&0.685		&	0.835&0.839		&	 \textbf{0.798}& \textbf{0.789}		\\
BPSQM- Pre		& 0.721&0.716		&	 \textbf{0.860}& \textbf{0.856}		& \textbf{0.776}& \textbf{0.773}		\\
BPSQM-Koniq10k	&  \textbf{0.756}& \textbf{0.734}		& \textbf{0.873}& \textbf{0.866}	&	- 	& - 			\\	
\hline \hline
\end{tabular}}
\end{table}

Fig. \ref{fig:nature} shows predicted quality maps describing authentic distortions on the CLIVE dataset. Specifically, in Fig. \ref{fig:nature} A(1), the human faces has obvious noises and the face regions in the quality map are darker than neighboring regions. The black clothing region is difficult for human vision to perceive prominent distortion effect, so the corresponding regions are brighter than other areas. In the Fig.  \ref{fig:nature} D(1), the face and clothes of the person on the left have evident blur distortion, so the output corresponding regions are close to dark. These cases suggest that the predicted quality maps of BPSQM are basically consistent with human judgments on authentic distortions.

A performance comparison of BPSQM and other methods on CLIVE is given in Table~\ref{table9}. It can be clearly observed that BPSQM performs better than most methods on the three authentic datasets. In particular, on CLIVE, BPSQM-Pre only obtains SRCC=0.716, PLCC=0.721 and does not achieve satisfactory metrics like those on synthetic distortion datasets. According to our analysis, the possible reasons are as follows. First, the proposed framework needs good intermediate quality maps to train the generative network, its overall performance mostly depends on the selected FR-IQA metric. Second, the diversity of training images content provided by synthetic dataset (CLIVE) is relatively simple with only 698 distorted images, that is to say the characteristics learned from generative network are not enough to cover the entire distribution of distortion in the image space, resulting to poor IQA prediction results. So we use the larger Koniq10k dataset to pre-train the entire model and then apply the pre-training parameters to initialize network, named as BPSQM-Koniq10k. As was expected, the prediction accuracy has significantly increased, which suggests that existing quality scored datasets are far from meeting the requirements of deep learning methods. This prompts us to pay more attentions to how to efficiently build a larger authentic distortion dataset in the future research. Interestingly and contrasting to the results on CLIVE and CID2013, on Koniq10k BPSQM-intial performs clearly better than BPSQM-Pre. The possible reason is that LIVE is a small dataset compared to Koniq10k in the richness of the image content, which does not have greater generalization performance than Koniq10k.

\subsection{Cross Dataset Test}\label{4.J}  

To evaluate the generalization performance of BPSQM, we utilize 80\% of the LIVE dataset for training and the remaining 20\% for verification, and the lowest validation loss model is selected to test on TID2008. we randomly used 80\% of the testing data for estimating parameters of a nonlinear functions.
\begin{equation}\label{q}
  \centering
  \tilde{q}=(\eta _{1}-\eta _{2})/(1+exp(-(\hat{q}-\eta _{3})/\left | \eta _{4} \right |))+\eta _{2}
\end{equation}
The reamining 20\% is used for testing. The median value of 100 random partition results is choosed as the final result.  Since the TID2008 includes more distortion types, we only chose the common types between the two databases, including JP2K, JPEG, WN and BLUR. Table~\ref{table10} shows that BPSQM performs well on the four distortion types. These results suggest that our proposed method does not depend on the database and shows good generalization capabilities.


\begin{table}
  \centering\caption{median SRCC results across 100 sessions on the TID2008 database}\label{table10}
  \begin{tabular}{c|c|ccccc} \hline \hline
     & Metrices    & JP2K  & JPEG  &   WN  &  BLUR & ALL \\ \hline
  FR & SSIM        & 0.963 & 0.935 & 0.817 & 0.960  & 0.902\\ \hline
  \multirow{3}{*}{NR}   & BRISQUE     & 0.832 & 0.924 & 0.829 & 0.881 & 0.896 \\
     & BIECON      & 0.878 & \textbf{0.941} & 0.842 & \textbf{0.913} & \textbf{0.923}\\
     & BPSQM-Fg-MD & \textbf{0.947} & 0.909 & \textbf{0.886} & 0.874  & 0.910 \\ \hline \hline
  \end{tabular}
\end{table}

\section{Conclusion} 

In this paper, we developed a simple yet effective blind predicting quality map for IQA that generates the map in pixel distortion level under the guidance of similarity maps derived by FR-IQA methods. Meanwhile, we also compare how to fuse the multi features information for predicting image quality. We believe that this proposed model could achieve better performance if where there is a better similarity index map navigating the generative network training. Optimizing our generative network to predict more accurate pixel distortion is a potential direction for future work.
\vspace{0.2cm}

\bibliographystyle{IEEEtran}   
\bibliography{conference_041818}   

\end{document}